\documentclass[11pt,a4paper]{article}
\usepackage[hyperref]{conll-2019}
\usepackage{times}
\usepackage{latexsym}
\usepackage{booktabs}
\usepackage{colortbl}
\usepackage{graphicx}
\usepackage{multirow}
\usepackage[skip=2pt]{caption}
\usepackage{textcomp}

\usepackage{url}

\aclfinalcopy 

\title{Diversify Your Datasets:\\Analyzing Generalization via Controlled Variance in Adversarial Datasets}

\author{Ohad Rozen$^1$, Vered Shwartz$^{2,3}$, Roee Aharoni$^1$, and Ido Dagan$^1$
       \\
       $^1$Computer Science Department, Bar-Ilan University, Ramat-Gan, Israel\\
       $^2$Allen Institute for Artificial Intelligence\\
       $^3$Paul G. Allen School of Computer Science \& Engineering, University of Washington\\
       {\tt\small \{ohadrozen,roee.aharoni\}@gmail.com, \tt\small vereds@allenai.org, \tt\small dagan@cs.biu.ac.il } \\
       }

\date{}

\begin{document}
\maketitle

\begin{abstract}
Phenomenon-specific ``adversarial'' datasets have been recently designed to perform targeted stress-tests for particular inference types. Recent work \cite{liu2019inoculation} proposed that such datasets can be utilized for training NLI and other types of models, often allowing to learn the phenomenon in focus and improve on the challenge dataset, indicating a ``blind spot'' in the original training data. Yet, although a model can improve in such a training process, it might still be vulnerable to other challenge datasets targeting the same phenomenon but drawn from a different distribution, such as having a different syntactic complexity level. In this work, we extend this method to drive conclusions about a model's ability to learn and \textit{generalize} a target phenomenon rather than to ``learn'' a dataset, by controlling additional aspects in the adversarial datasets. We demonstrate our approach on two inference phenomena -- dative alternation and numerical reasoning, elaborating, and in some cases contradicting, the results of \citeauthor{liu2019inoculation}. Our methodology enables building better challenge datasets for creating more robust models, and may yield better model understanding and subsequent overarching improvements.

\end{abstract}

\section{Introduction}
\label{sec:intro}
To successfully recognize textual entailment \cite[RTE; ][]{dagan2013recognizing}, also known as natural language inference (NLI) \cite{MacCartney:2008:MSC:1599081.1599147, bowman-etal-2015-large}, a system needs to model a broad range of inference phenomena. Pre-neural systems often included explicit components, such as engineered features or syntax-based transformations \cite[e.g.][]{Stern-2012-biutee, Stern:2012:EST:2390524.2390564, Bar-Haim:2015:KTI:2910557.2910558}, to address particular inference types such as syntactic, lexical, and logical inferences. Today's neural models do not explicitly model such inferences, but instead attempt to learn them implicitly from the training data. Despite their success on common NLI dataset, recent challenge datasets designed for probing different linguistic phenomena showed that neural models often fail on particular inference types, like recognizing semantic relations and negation \cite{poliak-etal-2018-collecting,naik-etal-2018-stress,glockner-etal-2018-breaking}. 

Recently, \newcite{liu2019inoculation} showed that when probing reveals a model's failure on a specific linguistic phenomenon, it is often possible to amend this failure. They suggested to fine-tune the model on (a training section of) the challenge dataset itself, in order to teach it to address the specific target phenomenon, or in other words, to ``inoculate'' it against the adversarial data. Inoculation has two possible outcomes. The first - a success to address the phenomenon after fine-tuning - suggests the original training set did not cover this phenomenon sufficiently (``blind spot'' of the dataset). A failure, on the other hand, indicates an inherent model weakness to handle the target phenomenon. This was presented as a general methodology, and was demonstrated on the NLI task, among others.

The inoculation approach seems an appealing way to teach NLI models to properly address a broad range of inference phenomena, by training on a targeted inoculation dataset for each phenomenon. However, the methodology as suggested in \citet{liu2019inoculation} is not conclusive as to whether inoculation succeeded thanks to the model learning the target phenomenon in a \textit{general} manner or due to overfitting the particular distribution of the inoculation data, possibly leveraging superficial cues or artifacts. Accordingly, successful inoculation may not reliably predict whether the model would successfully address the same target phenomenon when facing it on datasets drawn from different distributions.


In this paper, we extend the inoculation methodology, analyzing the ability of models to generalize across different data distributions when addressing a specific inference phenomenon. In particular, we suggest varying both training and test distributions along several linguistic dimensions, like syntactic complexity or lexical diversity. In addition to indicating the model generalization ability, our methodology directs how to design the inoculation data in order to sufficiently cover the targeted phenomenon, when possible.


We demonstrate our methodology on two inference types, picked from the GLUE benchmark  \cite{wang2018glue} diagnostic dataset: (1) dative alternation, a syntactic phenomenon, and (2) a specific type of numerical reasoning, pertaining to logical and arithmetic inference. To create our datasets, we introduce a templating method, by which we generated hundreds of synthetic examples from a single original sentence, while controlling the variance between the datasets.\footnote{All datasets and resources are available at https://github.com/ohadrozen/generalization.} We employ a recent NLI model, based on the pre-trained BERT masked language model \cite{devlin2018bert} fine-tuned on the MultiNLI dataset \cite{Williams2017Multi-NLI}. For the dative alternation case, we find that the model struggles with generalizing over the syntactic dimension, requiring training over a relatively large variety of syntactically complex sentences. For the numerical reasoning case, we find that the model notably fails to generalize across diverse number ranges, a conclusion that might have been missed if we were to use only the original inoculation methodology. We hope our methodology will be adopted for additional NLP tasks, and specifically to a broader range of entailment inference types, as an avenue for developing robust NLI systems that can address specific inference phenomena.

\section{Background}
\label{sec:background}
\paragraph{Neural NLI Models.} Natural language inference is the task of identifying, given two text fragments, whether the second (\emph{hypothesis}) can be inferred from the first (\emph{premise}). While earlier models for these tasks relied on domain knowledge and lexical resources like WordNet \cite[e.g][]{MacCartney:2008:PAM:1613715.1613817,Heilman:2010:TEM:1857999.1858143}, the release of the large-scale Stanford natural language inference dataset \cite[SNLI;][]{bowman-etal-2015-large} shifted the focus to neural models which thrive given such large datasets. Typically, these models encode each of the premise and the hypothesis, combine them into a feature vector, and feed it into a classifier to make the entailment prediction. The encoding of the two sentences can be either independent of each other or dependent using an attention mechanism. These models typically do not rely on any external knowledge other than pre-trained word embeddings. 

\paragraph{Contextualized Word Embeddings.} Recently, the word embedding paradigm shifted from static token-based embeddings to dynamic context-sensitive ones. Notable contextual representations are ELMo \cite{Peters2018Elmo}, BERT \cite{devlin2018bert}, GPT \cite{radford2018improving} and XLNet \cite{DBLP:journals/corr/abs-1906-08237}, which are pre-trained as language models on large corpora. Contextualized word embeddings have been used across a broad range of NLP tasks, outperforming the previous state-of-the-art models. Specifically, several works showed that they capture various types of linguistic knowledge, from syntactic to semantic and discourse relations \cite[e.g.][]{peters-EtAl:2018:N18-1,tenney2018what,lexcomp_tacl_2019}. Among many other tasks, NLI has also benefited from the use of contextualized word embeddings. The current state-of-the-art models use pre-trained contextualized word embeddings as their underlying representations, while fine-tuning on the NLI task \cite{Liu2019MNLI-SOTA, devlin2018bert}. Despite their remarkable success on several datasets, it still remains unclear how these models represent the various linguistic phenomena required for solving the NLI tasks.

\paragraph{Existing Drawbacks and Challenge Datasets.} Training an NLI model in this end-to-end manner assumes that any inference type involved in the sentence-level decision may be learned from the training data. However, recent work created challenge datasets which show that these models---when trained on the original NLI datasets---fail when they need to make inferences pertaining to certain linguistic phenomena, often ones which are not sufficiently represented in the training data. In these challenge datasets, a model is trained on the general NLI datasets, i.e. SNLI or the Multi-Genre Natural Language Inference datasets \cite[MultiNLI;][]{Williams2017Multi-NLI}. It is then used as a black box to evaluate on a given test set. 

\newcite{glockner-etal-2018-breaking} showed that substituting a single premise term with its hypernym (to create entailment examples) or a mutually exclusive term (for contradiction examples) challenges several pre-trained NLI models that performed well on the datasets on which they were trained. \newcite{naik-etal-2018-stress} constructed a suite of ``stress-tests'', each pertaining to some linguistic phenomenon, and showed that NLI models fail on many of them (e.g. numerical reasoning, logical negations, etc.). Another line of work showed that NLI models may  reach a surprising performance level on the NLI test sets just by exploiting artifacts in the generation of the hypotheses, rather than learning to model the complex entailment relationship \cite{gururangan2018annotation,tsuchiya2018performance,poliak-etal-2018-collecting}.

\paragraph{Fine-tuning on Challenge Datasets.} Recently, \newcite{liu2019inoculation} suggested that a model's failure to address a specific linguistic phenomenon may be attributed to one of the following cases: either the NLI training data does not sufficiently represent this phenomenon (``\emph{dataset blind spot}'') or the model is inherently incapable of learning to address this phenomenon. They suggested to fine-tune the NLI model on the specific challenge dataset in order to find out which case is currently observed. Specifically, in the case of a data blind spot, the performance on the challenge dataset is expected to improve after fine-tuning (i.e., the model is ``inoculated'' against the adversarial data). Otherwise, if the performance does not improve despite exposure to the phenomenon by fine-tuning, this may be an inherent weakness of the model. Finally, an additional possible outcome is that the performance on the original NLI test set is severely hurt after fine-tuning on the specific phenomenon, which may be due to over-fitting.

\section{Methodology}
\label{sec:methodology}
\begin{table*}[!t]
    \centering
    \small
    \begin{tabular}{p{0.2cm}ll}
    \toprule
    \rowcolor{lightgray} \multicolumn{3}{c}{Dative Alternation} \\ 
    (1) & \textbf{Extracted Premise:} & \textit{[Even our noble Saudi allies] [aren't willing to] lend [us] [their air bases].} \\
    (2) & \textbf{Premise Template:} & \textit{ARG$_1$ ARG$_2$ lend \textbf{ARG$_3$} ARG$_4$.} \\
    (3) & \textbf{Hypothesis Template (Ent. \#1):} & \textit{ARG$_1$ ARG$_2$ lend ARG$_4$ \textbf{to ARG$_3$}.} \\ 
    (4) & \textbf{Gen. Premise:} & \textit{[The allies across the sea] [have promised to] lend [\textbf{Italy}] [\underline{some of their land}].} \\
    (5) & \textbf{Gen. Hypothesis (Ent. \#1):} & \textit{The allies across the sea have promised to lend some of their land \textbf{to Italy}.} \\
    (6) & \textbf{Gen. Hypothesis (Ent. \#2):} & \textit{The allies across the sea have promised to lend \underline{some of their land}.} \\ 
    (7) & \textbf{Gen. Hypothesis (Cont.):} & \textit{The allies across the sea have promised to lend \textbf{Italy}}. \\
    \midrule
    \rowcolor{lightgray} \multicolumn{3}{c}{Numerical Reasoning} \\
    (8) & \textbf{Extracted Premise:} & \textit{[The Citigroup deal], [from beginning to end], [took] \textbf{less than} \underline{5} [weeks].} \\
    (9) & \textbf{Premise Template:} & \textit{ARG$_1$, ARG$_2$, ARG$_3$ \textbf{REL$_p$} \underline{NUM$_p$} ARG$_4$.} \\
    (10) & \textbf{Hypothesis Template (Ent.):} & \textit{ARG$_1$, ARG$_2$, ARG$_3$ \textbf{more than} \underline{NUM$_{smaller}$} ARG$_4$.} \\    
    (11) & \textbf{Gen. Premise:} & \textit{[My marriage], [despite much frustration], [lasted] \textbf{more than} \underline{7} [years].} \\
    (12) & \textbf{Gen. Hypothesis (Ent.):} & \textit{[My marriage], [despite much frustration], [lasted] \textbf{more than} \underline{2} [years].} \\
    (13) & \textbf{Gen. Hypothesis (Cont.):} & \textit{[My marriage], [despite much frustration], [lasted] \textbf{less than} \underline{5} [years].} \\
    (14) & \textbf{Gen. Hypothesis (Neutral):} & \textit{[My marriage], [despite much frustration], [lasted] \underline{8} [years].} \\
    \bottomrule
\end{tabular}

    \caption{Examples for the premise-hypothesis generation process (notations are explained in Sections \ref{sec:gen_premises} and \ref{sec:hypo_generation}): (a) Premises are extracted from the MultiNLI train set (rows 1 and 8) (b) Premise templates are manually created (rows 2 and 9) (c) Hypothesis templates are automatically generated using the premise templates (rows 3 and 10) (d) New premises are automatically generated by instantiating them with the turkers' answers (rows 4 and 11) (e) new hypotheses with same instantiations are generated (rows 5-7 and 12-14).}
    \label{tab:examples}
\end{table*}


We extend the inoculation approach of \citet{liu2019inoculation} by additionally controlling for finer-grained dimensions of the training and test data. For a given inference type (Section \ref{sec:inference_types}), our methodology consists of the following steps. (1) First, we extract \textit{premises} in the MultiNLI \cite{Williams2017Multi-NLI} training set that include the targeted linguistic phenomenon (Section~\ref{sec:find_premises}). (2) For each found premise, we generate multiple diverse variations using our templating method (Section \ref{sec:gen_premises}).
(3) After generating diverse premises, we generate multiple matching \emph{synthetic hypotheses} using a templating method (Section~\ref{sec:hypo_generation}). As we generate synthetic hypotheses, we can make sure the premise-hypothesis pairs differ along our proposed diversity dimensions. (4) Finally, we define the train and test sets so that the variance between them is controlled with respect to the different dimensions (Section~\ref{sec:train_test_split}). This facilitates probing the success of the model to generalize a given inference type with respect to a specific dimension of the data.

\subsection{Inference Types}
\label{sec:inference_types}

We focus on the following two inference types from the diagnostic set of
the GLUE benchmark \cite{wang2018glue} as test cases for our methodology. 
\paragraph{Dative Alternation.} This inference type refers to the alternation between a double-object construction (``\textit{I baked him a cake}'') and a prepositional indirect-object construction (``\textit{I baked a cake for him}''). 

\paragraph{Numerical Reasoning.} We focus on sentences relating to numbers by the relational phrases ``\textit{more than}'' and ``\textit{less than}'', normalizing all numbers to numerals. 
For example, ``\textit{There are \textbf{3} apples on the table}'' entails ``\textit{There are  \textbf{more than 2} apples on the table}''

\subsection{Premise Extraction}
\label{sec:find_premises}

For a given inference type, we start by finding premises in the MultiNLI train set that include the targeted linguistic phenomenon. We then construct templates based on these premises which are later used to generate synthetic premises and hypotheses. We do so in a semi-automatic way: we first use simple heuristics to track good candidates, and then manually select those that can be used as premises of NLI pairs that include the linguistic phenomenon in focus. For example, to track premises for the numerical reasoning inference type, we search for premises containing numbers and then choose the ones in which adding \textit{more than} or \textit{less than} before the number would keep the premise grammatical and coherent (e.g. ``\textit{the U.S. economy added \textbf{45} million jobs.}''), or ones which already include these terms before the number (e.g. ``\textit{The Citigroup deal, from beginning to end, took \textbf{less than 5} weeks.}''). We also make sure that we have enough diversity in the premise length and syntactic complexity (see Section~\ref{sec:train_test_split}). Rows 1 and 8 in Table~\ref{tab:examples} exemplify such premises for the dative alternation and numerical reasoning inference types. 

\subsection{Premise Generation}
\label{sec:gen_premises}
To isolate the lexical dimension from the syntactic one, for each target phenomenon we synthesize multiple new premises, all sharing a similar syntactic structure by construction. To do so, we manually generate a \emph{premise template} by replacing at least four spans in the premise with arguments ARG$_i$ as placeholders. We do so while keeping the words related to the phenomenon in focus within the template. For example, from the premise ``\textit{Even our noble Saudi allies aren't willing to lend us their air bases.}'',  we generated the template ``ARG$_1$ ARG$_2$ \textit{lend} ARG$_3$ ARG$_4$.'' (see rows 2 and 9 in Table~\ref{tab:examples}). We then let crowdsourcing workers instantiate each of the arguments to create new sentences. For the instantiations to later construct coherent sentences with high likelihood, we ask the workers to instantiate each argument separately, leaving the rest of the sentence unchanged, in a way that yields a new grammatical and coherent sentence that can make sense in some possible made up context (e.g. ``\textit{Even our noble Saudi allies \textbf{[span to fill in]} lend us their air bases.}''). To maintain similar sentence lengths and structures, we limit the instantiations to be at most one word longer or shorter than the original spans. For each argument we collected 6 instantiations which were manually validated for grammaticality and semantic coherence. We used all possible combinations of instantiations to generate hundreds of premises per template  (rows 4 and 11 in Table~\ref{tab:examples}). The annotation task was performed in Amazon Mechanical Turk, where to control the quality of the workers, we required that they have at least 98\% acceptance rate for prior HITs. We paid \$2.5 for two instantiations of all arguments in a given sentence (at most 14 instantiations per sentence).

\subsection{Hypothesis Generation}
\label{sec:hypo_generation}

For each \emph{premise template} we automatically generate multiple \emph{hypothesis templates} for each entailment label (entailment, neutral and contradiction), which differ from the premise only in the phenomenon in focus, as detailed for each inference type below.

\paragraph{Dative Alternation.} We generate the entailed hypothesis templates by applying the inference type. Specifically, for entailing dative alternation, we either switch to the alternate constructions (row 5 in Table~\ref{tab:examples}) or remove the first argument after the dative verb (row 6). For contradicting hypotheses templates, we remove the second argument, creating a grammatical yet contradictory hypothesis template (row 7). For each premise template we therefore generate 2 entailment hypotheses and 1 contradictory (no neutral).

\paragraph{Numerical Reasoning.} For a premise consisting of a target
numeric value NUM$_p$ preceded by a relational expression 
REL$_p \in\{less\ than, more\ than,\ \emptyset \}$, we generate multiple
hypotheses by replacing the target number by a random number (from a given target range, see further below) and the relational expression by each of the expressions. The sentence-pair label is determined by the relation between the numeric expressions in the premise and the hypothesis, and may be any of entailment, neutral, or contradiction. For example, consider the premise template ``[\textit{The union}] [\textit{has}] more than 4 [\textit{thousand members}] [\textit{in Canada}]''. Replacing the numeric expression ``\textit{more than 4}'' by ``3'' yields contradiction, while the substitute ``\textit{more than 3}'' is entailing, and ``\textit{more than 5}'' is neutral. This way, for each premise template we generate 22 different hypotheses: 4 entailment, 6 neutral and 12 contradiction. We used numeric values within the range 2-999.

\subsection{Controlled Data Splits}
\label{sec:train_test_split}
The main motivation for our data splits is to control the variance between the train and test sets with respect to certain data dimensions. Specifically, for both inference types, we create a variance along the syntactic complexity dimension. Based on the premise template, we divided each dataset into 3 subsets with different syntactic complexity levels: \emph{simple}, \emph{medium} and \emph{complex}, denoted by $S$, $M$ and $C$ respectively. We do so according to two criteria: sentence length and the depth in the constituency parsing tree in which the inference type occurs, using the Stanford Parser \cite{manning-EtAl:2014:P14-5} (see Table \ref{tab:data_stats}).\footnote{For dative alternation we consider the depth of the dative verb, while for the numeric reasoning data we look at the depth of the number. We consider premise templates with less than 16 words and depth $<4$ as \textit{simple}, more than 25 words and depth $>6$ as \textit{complex}, and the rest as \textit{medium}.} 

We also create a variance along different lexical dimensions. From the \emph{simple} subset $S$ we generate two additional subsets $S_{Lex1}$ and $S_{Lex2}$ in the following way. For each \textit{simple} template, we first split its original instantiations into two groups, and then instantiate each such group separately into the template, creating two sets of instantiations of the same template $s_1 \in\ S_{Lex1}$ and $s_2 \in\ S_{Lex2}$, which are syntactically similar by construction, yet lexically different. We repeat this process for the \emph{complex} subset as well to create $C_{Lex1}$ and $C_{Lex2}$. We further split the dative alternation datasets lexically by the main verb, which allows testing how well the model generalizes this inference type for different dative verbs. This is done by changing the main dative verb in $S_{Lex2}$ and $C_{Lex2}$ creating new subsets $S'_{Lex2}$ and $C'_{Lex2}$.\footnote{For each template we choose a new dative verb that keeps the sentence coherent and grammatical.} We split the numerical reasoning dataset similarly using different numerical ranges in the training and test sets.

In Section~\ref{sec:experiments} we experiment with various splits of the training and test sets, which allow us to test the model's generalization over the various dimensions. For example, we test whether the model can learn an inference type on syntactically \emph{simple} examples and generalize it to \emph{complex} ones. See Table \ref{tab:data_stats} for the statistics of each dataset. 

\begin{table}[!t]
    \centering
    \small
    
\begin{tabular}{p{1.4cm}ccccc}
    \toprule
    \textbf{} & \textbf{Simple} & \textbf{Medium} & \textbf{Complex} & \textbf{All} \\
    \midrule
    \rowcolor{lightgray} \multicolumn{5}{c}{Number of Premise Templates} \\ 
    \textbf{Datives} & 10 & 9 & 9 & 28 \\
    \textbf{Numbers} & 9 & 12 & 9 & 30 \\
    \rowcolor{lightgray} \multicolumn{5}{c}{Number of Examples} \\ 
    \textbf{Datives} & 21K & 36K & 34K & 91K  \\
    \textbf{Numbers} & 181K & 239K & 182K & 602K  \\    
    \bottomrule
\end{tabular}


    \caption{Statistics of the dative alternation and numerical reasoning datasets divided by syntactic complexity level.}
    \label{tab:data_stats}
\end{table}

\section{Experiments}
\label{sec:experiments}

We use a standard model based on BERT \cite{devlin2018bert} as our NLI model. Specifically, we used the base-uncased pre-trained model from the pytorch-pretrained-bert library\footnote{https://github.com/huggingface/pytorch-pretrained-bert}, and fine-tuned it for the NLI task on MultiNLI.\footnote{The MultiNLI dataset has domain-matched and mismatched development data. We use ``matched'' for our testing.} We conduct several experiments. First, we use our datasets for a typical probing task, i.e. testing how well the model performs on each inference type, \emph{without being trained to address the specific inference type} (Section~\ref{sec:probing}). Then, similarly to \newcite{liu2019inoculation}, we test the model's ability to learn each inference type by \emph{further fine-tuning on specific examples for it} (Section~\ref{sec:fine_tune}). Finally, incorporating our innovation, we analyze the model's generalization ability by introducing variance in the proposed data dimensions between the train and test sets (Section~\ref{sec:generalization}).

\begin{table}
    \centering
    \small
    

\begin{tabular}{lcccc}
    \toprule
    \textbf{Complexity} & \textbf{Ent.} & \textbf{Neutral} & \textbf{Cont.} & \textbf{All} \\
    \midrule
    \rowcolor{lightgray} \multicolumn{5}{c}{MultiNLI Matched Dev Set} \\     
    \textbf{All} & 83.56 & 84.12 & 86.37 & \textbf{84.66} \\
    \rowcolor{lightgray} \multicolumn{5}{c}{Dative Alternation} \\ 
    \textbf{Simple} & 100 & - & 4.22 &  \textbf{52.63}  \\
    \textbf{Medium} & 100 & - & 2.16 & \textbf{49.27} \\
    \textbf{Complex} & 99.77 & - & 0.36 & \textbf{50.45} \\
    \textbf{All} & 99.92 & - & 2.25 & \textbf{50.78} \\
    \midrule
    \rowcolor{lightgray} \multicolumn{5}{c}{Numerical Reasoning} \\ 
    \textbf{Simple} & 38.14 & 0.66 & 69.53 & \textbf{45.04}  \\
    \textbf{Medium} & 57.14 & 1.36 & 50.14 & \textbf{38.11}  \\
    \textbf{Complex} & 55.48 & 3.04 & 46.26 & \textbf{36.15} \\
    \textbf{All} & 50.25 & 1.69 & 55.31 & \textbf{39.77} \\
    \bottomrule
\end{tabular}

    \caption{Accuracy of the model trained only on MultiNLI on our datasets, which are used as probing datasets. The Complexity column refers to the syntactic complexity of the sentences.}
    \label{tab:probing_results}
\end{table}

\begin{figure*}[!t]
    \centering
    \begin{tabular}{cc}
        \includegraphics[width=0.45\textwidth]{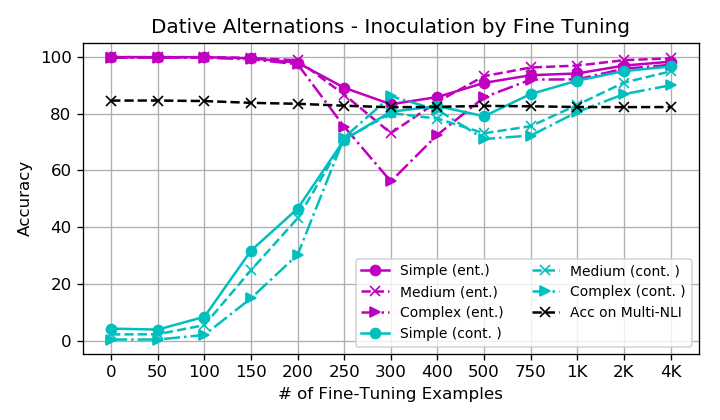}
        &
        \includegraphics[width=0.45\textwidth]{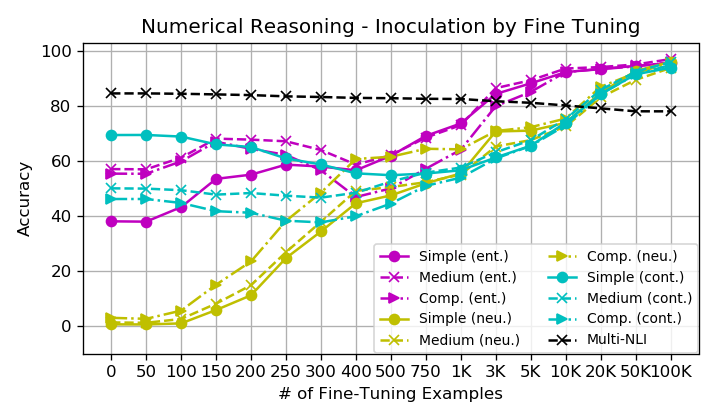}
                  \\
    \end{tabular}
    \caption{Average test accuracy on the dative alternation dataset (left) and the numerical reasoning dataset (right) as a function of number of training examples, divided by label. The black lines represent the average accuracy of the model on MultiNLI matched development set after fine-tuning. For numerical reasoning we use a larger range on the x-axis to capture the near-perfect performance for 100k examples.}
    \label{fig:fine_tuning}
\end{figure*}

\subsection{Probing}
\label{sec:probing}

We randomly selected 4,000 examples from each of the \textit{simple}, \textit{medium} and \textit{complex} datasets, with balanced labels, to serve as a test set. Table~\ref{tab:probing_results} shows the accuracy of the model on each test set. 

On our dative alternation dataset, the accuracy on our test sets is substantially lower than on the MultiNLI development set (50.78\% versus 84.66\% respectively), suggesting that the model has not learned to address this inference type from the MultiNLI training data. The model has very high accuracy on the entailment examples, while close to zero on the contradiction ones. This is understandable considering that the sentence-pairs in this dataset by construction have high lexical overlap between the premise and hypothesis, leading the model to default to almost always predicting entailment.

On the numerical reasoning dataset, the model also seems to fail on this inference type with test set accuracy much lower than on the MultiNLI development set, suggesting that the model hasn't learned to address this inference type as well. The model has relatively low accuracy on the entailment and contradiction examples while close to zero accuracy on the neutral ones. This is due to the 
fact that the model classifies sentence-pairs with high lexical overlap but with a different numerical phrase as either entailment or contradiction, but almost never as neutral.

\subsection{Fine Tuning}
\label{sec:fine_tune}

We follow \newcite{liu2019inoculation} and fine-tune the model on the phenomenon-specific examples, testing how many training examples the model needs to observe before it performs reasonably well on this inference type.\footnote{To fine-tune BERT, we use the Adam optimizer \cite{DBLP:journals/corr/KingmaB14} with a learning rate of $7\cdot10^{-7}$,  $\beta_1 = 0.9$, $\beta_2 = 0.999$, and $L_2$ weight decay of $0.01$.}

We split each dataset to training (77\%) and test (23\%) sets such that the same template is not used in both training and test. Each set consists of templates from all syntactic complexities, and a balanced number of examples from each label. We experiment with a different number of training examples ranging from 0 to 4,000 for the dative alternation sets and from 0 to 100,000 for numerical reasoning. We repeat this experiment five times with different  training and test splits, while also testing the performance on the original MultiNLI matched development dataset. We report in Figure~\ref{fig:fine_tuning} the average accuracy across runs as a function of the number of training examples. In both datasets, fine-tuning greatly improves the performance.

On the dative alternation data, fine-tuning brings the performance on contradiction from 0 to 90\%, suggesting the model can now distinguish well between the entailing and contradictory examples, reaching similar accuracy on both. The good performance on this inference type indicates a blind spot in the MultiNLI dataset rather than a model weakness. As expected, the model reaches slightly better performance on examples with simpler syntactic structure than those with complex ones.  Fine-tuning with 4,000 examples reduces the performance on the MultiNLI development set in about 2\%. As \citeauthor{liu2019inoculation} suggested, this could result from the distribution of our dataset deviating from the distribution of the MultiNLI dataset, and possibly from having the original model overfitting to that distribution.

With respect to numerical reasoning, though only after a relatively large number of 10,000 training examples, the model seems to succeed in learning the phenomenon. According to \newcite{liu2019inoculation}, this result suggests that our challenge dataset did not reveal a weakness in the model, but instead a blind spot in the original dataset. Yet, this conclusion is challenged in the next subsection. In the numerical reasoning case we notice an even larger decrease in the performance on  MultiNLI after fine-tuning - up to 6.5\%. This may again result from different distributions across the datasets, influencing the performance more intensely due to the larger number of training examples.

\begin{figure*}[!t]
    \centering
    \includegraphics[width=0.45\textwidth]{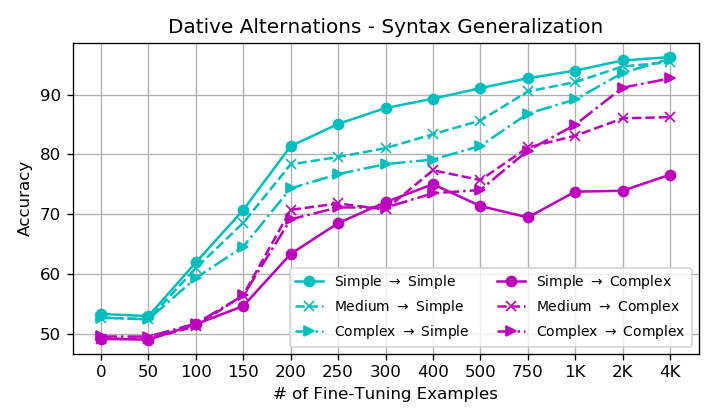}~
    \includegraphics[width=0.45\textwidth]{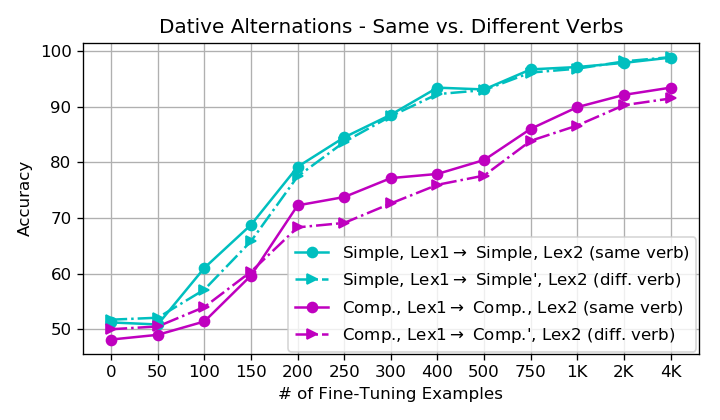}
    \caption{Test accuracy on the dative alternation dataset over the syntactic dimension (left) and average performance on the lexical dimension (right) as a function of number of training examples. The larger gaps on the left graph in  comparison to the much smaller gaps on the right graph indicate a limited generalization ability over the syntactic dimension and a better one over different dative verbs}
    \label{fig:generalization_datives}
\end{figure*}

\begin{figure*}[!t]
    \centering
    \includegraphics[width=0.45\textwidth]{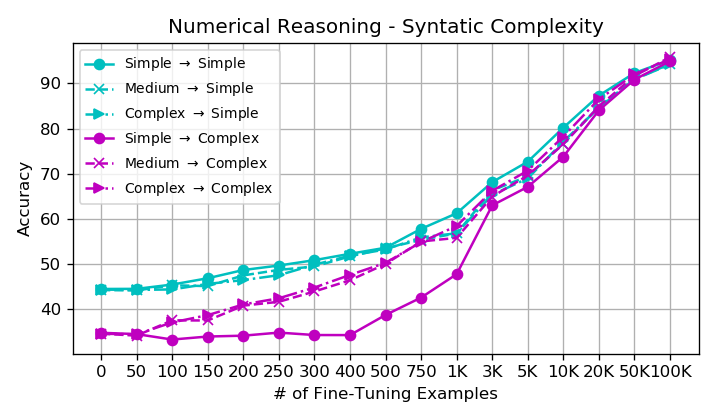}~
    \includegraphics[width=0.45\textwidth]{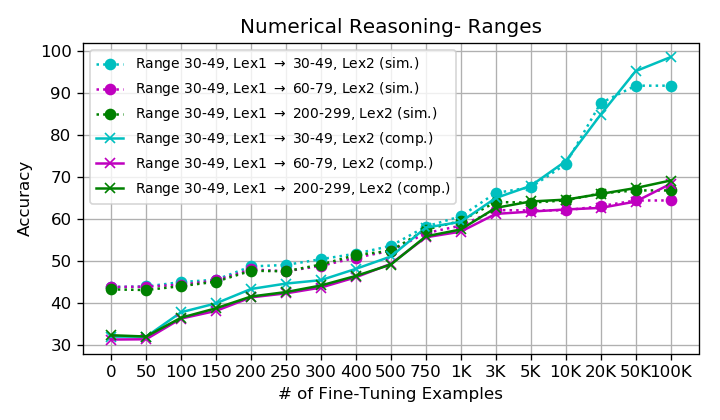}
    \caption{Test accuracy on the numerical reasoning dataset over the syntactic dimension (left) and on the range dimension (right) as a function of number of training examples. The gaps on the right graph comparing to the convergence on the left one suggest a good generalization ability over the syntactic dimension and a poor one over different number ranges.}
    \label{fig:generalization_numbers}
\end{figure*}

\subsection{Generalization}
\label{sec:generalization}

We now analyze the model's ability to generalize for each inference type across various data dimensions, using our proposed methodology. As we will see, this type of analysis yields additional, more elaborate and sometimes contradictory insights, which are not attainable by the prior methodologies that we applied in the previous two subsections.

\paragraph{Dative Alternation.} First, we test the model's generalization ability at the syntactic complexity dimension, by training it on a dataset belonging to one category of syntactic complexity (among \textit{simple}, \textit{medium} and \textit{complex}; see Section~\ref{sec:train_test_split}), and testing it on a dataset belonging to either  \textit{simple} or \textit{complex}. We make sure examples in the training and test sets were generated by different templates.

The left side of Figure~\ref{fig:generalization_datives} displays the performance on the various experiments, revealing an interesting pattern: on the \textit{simple} syntax test set, good performance is attainable regardless of the training set, while for the \textit{complex} syntax test set, training on simple syntax performs inferiorly to training on complex syntax. This suggests that although the model is able to learn the dative alternation phenomenon and generalize to a certain extent, the model does not learn the phenomenon on its own, decoupled from learning argument positions, but rather it needs to be trained with dative alternation examples of high syntactic complexity to perform well.


The second data dimension we test is lexical diversity. We test the model's ability to generalize across syntactically-similar examples with a different main dative verb. To isolate the lexical aspect from other aspects, we fix the syntax by using the same templates for training and testing. For the \textit{simple} category, we train the model on the $S_{Lex1}$ subset and test it on both the $S_{Lex2}$ subset with the same main dative verb, and on $S'_{Lex2}$ with a different main verb that has not been seen in the training examples (see Section~\ref{sec:train_test_split}). We repeat the same process for the \textit{complex} category.\footnote{For both $S_{Lex1}$ and $C_{Lex1}$ we sample 256 examples from each of 5 manually chosen templates from the related category.}

The right side of Figure~\ref{fig:generalization_datives} shows that when tested on the same syntactic complexity level as seen during training, the performance remains similar regardless of the similarity between the train and test dative verbs. This suggests that the model generalizes well on the lexical dimension and learns to recognize the dative alternation inference independently of the specific verb. We also observe that the model generalizes more easily from examples with simple syntax: on this category, the performance gap between the two test sets $S_{Lex2}$ and $S'_{Lex2}$ is smaller than the gap between the graphs of the complex category. This suggests the conclusion that unlike syntactic diversity, a large lexical diversity is not necessary when inoculating for dative alternation. 

\paragraph{Numerical Reasoning.} Again, we test the model's generalization ability with respect to syntactic complexity by splitting the train and test sets based on this dimension (left side of Figure~\ref{fig:generalization_numbers}). As opposed to dative alternation, here the gap between the performance when testing examples with more complex syntax than the training set and the performance when testing on simpler examples is rather small after enough training examples (up to 3.2\% difference after 3,000 examples). We conjecture that the model learns to identify local patterns (e.g. ``\textit{more than} X'') while the relationships between them are less dependent on the global syntactic structure of the sentence. Moreover, the difference between the premise and the hypothesis is lexical and local (i.e. replacing the relational operator and the number), while their syntax remains otherwise identical, shifting the model's focus away from the syntax. 

Regarding lexical diversity, we test whether the model can be trained on one range of numbers and perform well when tested on another. To that end, we populate the number placeholders in the templates with randomly sampled numbers from within a certain range, among 30-49, 60-79 and 200-299. We train on the first range and test on each of the ranges. To isolate the numeric aspect from the syntactic one, we fix the syntax by using the same templates for training and testing, while using different argument instantiations (as we did for the dative alternation when testing for lexical diversity of the dative verb).

The right side of Figure~\ref{fig:generalization_numbers} shows the performance of each model as a function of training examples. Training and testing on the same range yields substantially better performance, while testing on a different range reaches accuracy of less than 70\% even after 20,000 training examples, and seems to reach saturation. We also repeated the same experiment with number range of 1000-9999 in the test sets, resulting in a graph very similar to the 200-299 range. This indicates an inherent weakness of the model to learn the phenomenon and generalize it over different number ranges. Given that the number of training examples is limited, it might be challenging to inoculate the model to perform well on a wide variety of challenge datasets for this phenomenon. The success within the same (narrow) number range, when training gets large enough, might suggest that the model mostly memorizes specific number pairs and the arithmetic relation between them, rather than learning the arithmetic rules. These insights contradict the conclusion of the original inoculation analysis of a blind spot in the original dataset (in Section~\ref{sec:fine_tune}).

\section{Conclusions}
\label{sec:conclusions}
We presented a methodology to analyze the ability of NLI models to learn a specific inference phenomenon and successfully generalize its modeling to datasets drawn from different distributions. By controlling the differences between the training and test sets along syntactic and lexical data dimensions, we were able to analyze how well the model generalizes with respect to each phenomenon over the different dimensions. We demonstrated our methodology on a standard model based on BERT, focusing on dative alternation and numerical reasoning. We found that high syntactic complexity is necessary for teaching dative alternations, while being less important for numerical reasoning. We also showed that the model is incapable of generalizing over different number ranges for numerical reasoning, indicating an inherent modeling weakness.

We suggest that our work opens promising as well as challenging research directions. A natural direction for future work would be to apply our methodology to a broader range of inference types and data dimensions. This would enable extensive analysis of NLI models' learning and generalization abilities, and may yield models that can truly address a range of inference phenomena in a fairly general manner. One question that still remains open at this point regards the models' ability to handle \textit{multiple} inference phenomena within the same example, which is more representative of real-world scenarios.

Finally, we observed that fine-tuning the model on the phenomenon-specific data sometimes yields a decrease in the performance on the original dataset. One potential direction to avoid this in the future would be to perform multi-task learning rather than fine-tuning on the phenomenon-specific data. Better scheduling of multi-task training as presented by \newcite{kiperwasser2018} may also reduce the performance loss in such scenarios.

\section*{Acknowledgments}
We would like to thank Ori Shapira for assisting in data analysis, and the anonymous reviewers for their constructive comments. This work was supported in part by the German Research Foundation through the German-Israeli Project Cooperation (DIP, grant DA 1600/1-1), by a grant from Reverso and Theo Hoffenberg, and by the Israel Science Foundation (grant 1951/17).

\bibliography{references}
\bibliographystyle{acl_natbib}

\end{document}